%% file: main.tex
\pdfoutput=1

\documentclass[11pt]{article}

\usepackage{acl}

\usepackage[normalem]{ulem} 
\usepackage{stackengine}

\usepackage{times}
\usepackage{latexsym}
\usepackage{tabularx}
\usepackage{amsmath}
\usepackage{booktabs}
\usepackage{multirow}
\usepackage[T1]{fontenc}
\newcommand{\llm}[1]{\textsc{#1}}

\usepackage[utf8]{inputenc}

\usepackage{microtype}
\usepackage{xcolor}
\usepackage{tcolorbox}
\usepackage{adjustbox}
\usepackage{tikz}
\usepackage{graphicx}
\usetikzlibrary{shapes.geometric, positioning, shapes.symbols, arrows.meta}

\definecolor{pastelRed}{rgb}{1.0, 0.8, 0.8}
\definecolor{pastelGreen}{rgb}{0.01, 0.75, 0.24}
\definecolor{bluebell}{rgb}{0.64, 0.64, 0.82}\definecolor{pastelYellow}{rgb}{1.0, 0.93, 0.61}
\definecolor{beaublue}{rgb}{0.74, 0.83, 0.9}
\definecolor{darkerbeaublue}{rgb}{0.64, 0.73, 0.8}
\definecolor{pastelGrey}{rgb}{0.83, 0.83, 0.83}
\definecolor{celadon}{rgb}{0.67, 0.88, 0.69}
\definecolor{bubblegum}{rgb}{0.99, 0.76, 0.8}
\definecolor{lightgray}{gray}{0.9}

\newcommand{\inlinecolorbox}[2]{
  \tikz[baseline=(X.base)] \node [rectangle, rounded corners=1mm, fill=#1, inner sep=0.5mm] (X) {#2};%
}

\usepackage{inconsolata}

\usepackage{graphicx}
\usetikzlibrary{shapes.geometric, arrows, positioning}

\tikzstyle{startstop} = [rectangle, rounded corners, minimum width=3cm, minimum height=0.8cm,text centered, draw=black, fill=gray!20]
\tikzstyle{process} = [rectangle, minimum width=3cm, minimum height=0.8cm, text centered, draw=black]
\tikzstyle{arrow} = [thick,->,>=stealth]
\tikzstyle{decision} = [diamond, minimum width=3cm, minimum height=0.8cm, text centered, draw=black, fill=white]
\tikzstyle{robot} = [rectangle, inner sep=0, minimum size=1cm]
\newcommand{\cut}[1]{}
\newcounter{tbsnr}
\newenvironment{tbs}
{\addtocounter{tbsnr}{1}\par\bigskip\noindent\fbox{\thetbsnr}
\hspace*{\fill}\begin{minipage}{7cm}\tt}
{\end{minipage}\hspace*{\fill}\bigskip}

\newcommand{\at}[1]{\textcolor{red}{\textbf{AT: #1}}}

\title{Learning to Ask Informative Questions:  Enhancing LLMs with \\ Preference Optimization and Expected Information Gain}

\author{Davide Mazzaccara \\
CIMeC, University of Trento\\
\texttt{davide.mazzaccara@unitn.it}\\\And
Alberto Testoni\\
ILLC, University of Amsterdam\\
\texttt{a.testoni@uva.nl}\\\AND
Raffaella Bernardi\\
CIMeC, DISI, University of Trento\\
\texttt{raffaella.bernardi@unitn.it}}

\begin{document}
\maketitle
\begin{abstract}

Questions are essential tools for acquiring the necessary information to complete information-seeking tasks. However, large language models (LLMs), especially open-source models, often perform poorly in generating informative questions, as measured by expected information gain (EIG). In this paper, we propose a method to enhance the informativeness of LLM-generated questions in 20-question game dialogues. We sample multiple questions from the same model (\llm{Llama 2-chat 7B}) for each game and create pairs of low-EIG and high-EIG questions to apply a Direct Preference Optimization (DPO) algorithm. Our results show that this method produces more effective questions (in terms of EIG), even in domains different from those used to train the DPO model.

\end{abstract}

\input{latex/introduction}

\input{latex/methodology}

\input{latex/results}
\input{latex/conclusion}
\input{latex/limitations}
\section*{Acknowledgments}
We are grateful to David Schlangen and his group and to the audience of the Evil seminar organized by Marco Baroni. By addressing their questions the quality of the work improved. We thank Leonardo Bertolazzi for his suggestions on the early phase of the implementation of this work. Alberto Testoni is supported by the European Research Council (ERC) funding under the European Union’s Horizon 2020 research and innovation programme (grant agreement No.~819455, PI R. Fernández).

\bibliography{custom.bib}

\nocite{Apple, Confirm-it, GroundedRetrieval23, Testoni23, STAR22, INLG23, Jorge2017learning}

\appendix
\input{latex/appendix}

\end{document}

%% file: latex/introduction.tex
\section{Introduction}
\label{sec:intro}



Questions in language serve as requests for information \citep{Searle1969SpeechActs}. A speaker lacks information in their knowledge state and asks questions to gain this information. This process of acquiring information through questioning is essential for children to learn about the world \citep{Chouinard2007} and for adults to solve complex problems \citep{Geva2021Aristotle}. 
Questions, however, vary in their level of informativeness, with some questions being more informative and efficient in reaching the problem's solution \citep{grand:LIPS}.

Cognitive Science provides two interesting tools to study questions' informativeness: the 20 Questions Game as a test bed, and the Expected Information Gain (EIG) as a quantifying measure \citep{RUGGERI2015203}. The 20 Questions game consists of one player asking yes/no questions to identify the item the other player has in mind, from a pool of possible items. In this context, the EIG \citep{shannon1948mathematical} measures questions' informativeness as the (expected) entropy reduction caused by a certain question in the space of possible items. Most informative polar questions partition the possible items into two same-size clusters: items in one cluster are expected to receive a positive answer, and items in the other cluster are expected to receive a negative one.



In recent years, Large Language Models (LLMs) have demonstrated remarkable language and reasoning capabilities \citep{kojima2022large, huang-chang-2023-towards}. The ability to ask informative and effective questions is crucial for employing these systems as successful user assistants on a large scale. Although LLMs have demonstrated their ability to play the 20 Questions game, their questions are characterised by low informativeness and limited success \citep{INLG23}. Recent studies have emerged to improve the informativeness of LLMs' questions. \citet{Apple} improve open-source LLMs via Reinforcement Learning and Behavioral Cloning from larger LLMs' data. Relying on LLMs' ability to generate diverse questions and provide reliable answers \citep{piriyakulkij2023activepreference, Testoni23}, \citet{hu:unce24} propose an inference-time re-ranking strategy based on EIG. Differently from them, we propose a training strategy leveraging EIG as a refined signal with a preference optimization algorithm. Unlike other approaches, all the steps of our method are accomplished by the same open-source model, without requiring annotation or feedback signals from external models.

\begin{figure*} 
\begin{center}
\includegraphics[width=1\linewidth]{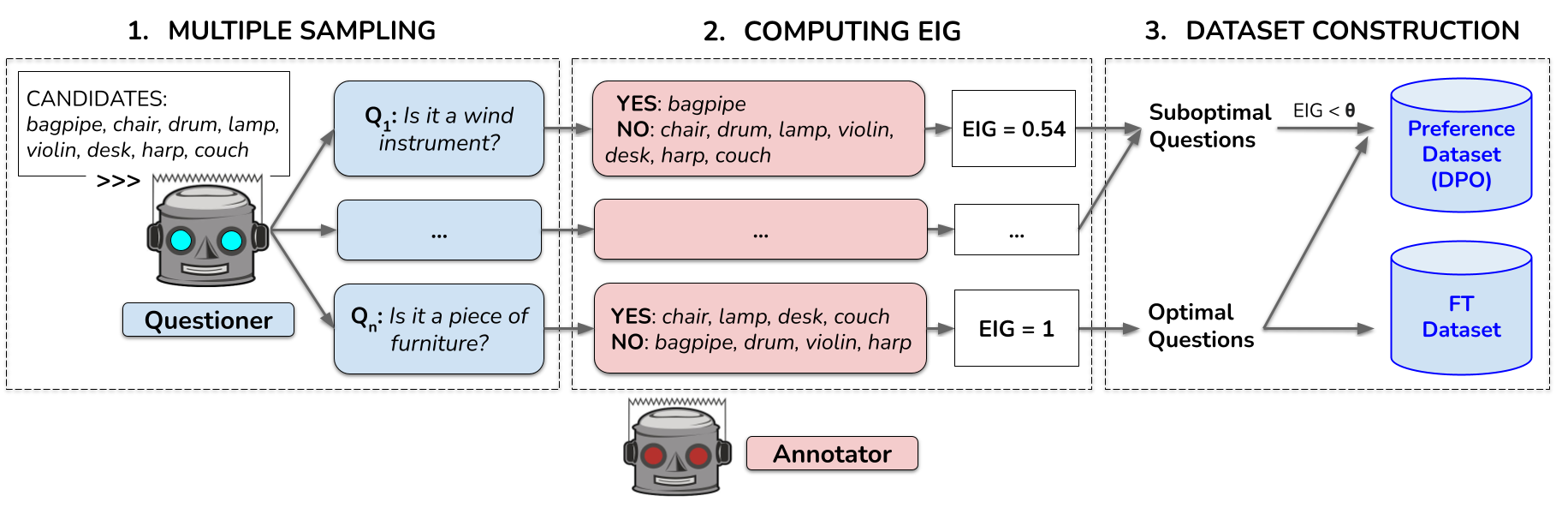}
\caption{The proposed approach for constructing the datasets of dialogues for fine-tuning and preference optimization (DPO). From the original candidate set, the Questioner generates a question Q$_1$, and the Annotator provides the expected answer to each candidate. Expected Information Gain (EIG) is computed from the annotation: if the question is suboptimal in terms of EIG, other questions are sampled until an optimal question is reached (Q$_n$). The optimal question is paired with the suboptimal ones in the Preference dataset (DPO), whereas the Fine-Tuning (FT) dataset is composed of only 1-EIG questions.} \label{fig:bootstrapping_new}
\end{center}
\end{figure*}

This study aims to improve the informativeness of questions generated by LLMs, thereby leading to more efficient agents. The 20 Questions Game serves as a testbed to illustrate the feasibility of our approach.
To achieve our aim, we propose a method involving three steps: sampling multiple questions, questions' evaluation in terms of EIG, and training with preference optimization.  In the first step, the model generates a set of possible questions, and it computes the EIG for each question. This set of questions, along with the corresponding EIG values, as a proxy for the questions' informativeness, is used for Direct Preference Optimization (DPO; \citealt{rafailov2024direct}). Our results show that EIG is a strong training signal to improve the question-asking capabilities of current LLMs and overcome their shortcomings in asking effective questions.\footnote{Code and data are available at \url{https://github.com/dmazzaccara/LearningToAsk}}

While these findings hold in the controlled setting of the 20 Questions Game, our approach could be extended outside this paradigm and metric to train different models’ capabilities. Our core idea of sampling, self-annotation to filter with a metric, and preference optimization could improve questions’ generations in a plethora of tasks – recommendation systems \citep{piriyakulkij2023activepreference} and image retrieval \citep{GroundedRetrieval23} for example. Other metrics, such as Expected Savings \citep{Rothe2018askgood}, could be employed to generate large preference datasets and improve LLMs' information-seeking abilities beyond our tested paradigm.

%% file: latex/methodology.tex
\section{Methodology}
\subsection{Setting} \label{Setting}


Our experimental setting, inspired by the 20 Questions Game paradigm, involves two players with different roles: a Questioner and an Answer. The Answer is secretly assigned one target entity $\omega$ among a pre-defined set of candidates (the \emph{candidate set} – $\Omega$). The Questioner receives the initial candidates and is instructed to ask yes/no questions
, in order to identify $\omega$.

Each \emph{game} consists of a set of possible items, $\Omega$, available to the Questioner, and a specific target item, $\omega$, available to the Answerer, where $\omega \in \Omega$. A \emph{dialogue} is a series of question-answer exchanges between the two players, ending when the Questioner correctly identifies the target or reaches the maximum number of questions (20). If the target is reached within the 20 questions, the dialogue is \emph{successful}; otherwise, it is \emph{unsuccessful} (see more details on the successful condition in Appendix \ref{appendix:successful_cond}). For a given game, if the dialogue is not successful on the first attempt, the Questioner and the Answerer can engage in up to $n$ dialogues to identify the target, where $n = |\Omega|/2$. 
\cut{A \emph{dialogue} is a series of question-answer exchanges between the two players, ending when the Questioner makes a guess about the target. If the guess is correct, the dialogue is considered \emph{successful}; otherwise, it is \emph{unsuccessful}.} 



\subsection{Data Sampling}
Following the procedure in~\citet{INLG23}, 10,000 games are extracted from McRae concepts \citep{mcrae2005norms}. The candidate sets consist of concepts belonging to 6 categories (mammal, bird, clothing, weapon, fruit, and vegetable) for a total of 66 unique concepts. Each candidate set consists of 8 elements ($|\Omega|=8$) belonging to two categories of 4 elements each. 

In our setting, all the game's roles are played by the same LLM, \llm{Llama 2-chat 7B} \cite{Llama2}. Along with the Questioner and the Answerer, sampling involves the Annotator, who provides the annotation to compute the EIG. As illustrated in Fig. \ref{fig:bootstrapping_new}, English dialogues are generated from each game by repeatedly sampling questions from the Questioner. For each generated question, the Annotator provides the yes/no annotation ($\Omega_{yes}$ and $\Omega_{no}$). If the question splits the candidate sets in two equal subsets ($|\Omega_{yes}| = |\Omega_{no}|$), the question is \emph{optimal} in terms of EIG (details on EIG computation are in Appendix \ref{appendix:eig}). Once an optimal question is reached, it is saved along with the answer by the Answerer in the dialogue history. This dialogue history is provided as input to the Questioner to sample the optimal follow-up question (implementation's details are in Appendix \ref{appendix:setting}). 

This process results in two clusters of optimal and suboptimal questions, from which two datasets are obtained. A \emph{fine-tuning dataset} consisting of dialogues of only optimal questions ($EIG=1$), indicating that they evenly split the candidate space at each turn. A \emph{preference dataset}, which consists of 55,000 optimal ($EIG = 1$) vs suboptimal ($EIG < 0.8$) question pairs.





\subsection{Questioner Models}

Our evaluation of the Questioner compares \llm{Llama 2-chat (7B)} zero-shot with the trained versions below. Both the training processes rely on LoRA adapters and standard hyperparameter settings.

\noindent \textbf{Fine-tuning} (FT) involves a causal language modelling objective applied to the entire sequence. For development, we use 50 games excluded from the fine-tuning dataset. To obtain the best adapter's checkpoint, 
we test the adapter on the development set every 1,000 training samples. According to the number of questions per dialogue, the best adapter's checkpoints are after 4,000 dialogues. \\
\noindent \textbf{Direct Preference Optimization} (DPO) \citep{rafailov2024direct} training objective increases the likelihood of optimal EIG questions while decreasing the likelihood of the low-EIG question. For DPO, we train with all the 55,000 pairs of optimal vs low-EIG questions.

\subsection{Evaluation}
\textbf{Test Sets}: three test sets evaluate questioner models with candidate sets of the same size as in training ($|\Omega|=8$): INLG, Things, Celebrities; two test sets test with more candidates than in training: INLG 16 and BigBench. The same-size test sets have candidate sets from different domains: INLG \citep{INLG23} from seen categories but unseen concepts; both from \citet{Apple}, Things has unseen categories but common-life concepts, Celebrities has unseen categories and concepts compared to training (famous personalities). Different-size test sets are INLG 16 \citep{INLG23}, with sets of $|\Omega|=16$ unseen concepts of seen categories, and BigBench, with sets of $|\Omega|=29$ different categories and concepts \citep{srivastava2023beyond}. Details are in Appendix~\ref{sec:apdataset}.

\noindent \textbf{Metrics}: the impact of FT and DPO on zero-shot is assessed with three metrics. Task success (\textbf{S@1}) is the percentage of games in which the model identifies the target within the first dialogue. Average number of questions (\textbf{AQ}) is the number of questions to reach the target in successful dialogues. Expected Information Gain (\textbf{EIG}) is the averaged questions' EIG in successful dialogues. 





%% file: latex/results.tex
\section{Results}\label{sec:results}
Different methods are assessed on test sets with candidate sets of the same size as in training but from other domains. The results are reported in Table~\ref{tab:8cds}.
For INLG 
and Things,
DPO significantly improves performance over the zero-shot model in terms of both S@1 ($+12.2\%$ and $+10\%$) and AQ ($-2.1$ and $-2.3$ questions per successful dialogues). For Celebrities,
where the concept space differs greatly from the training data, DPO reduces the average number of questions by $2.5$, while S@1 shows only a marginal improvement. A consistent improvement of DPO over zero-shot is also observed for the average EIG of the generated questions. Overall, the fine-tuning approach leads to a significant degradation in the S@1 metric. 
\begin{table}[h]
\centering
\resizebox{1.00\linewidth}{!}{
\begin{tabular}{llccc} 
\toprule
Set & Method & S@1 \scriptsize$\uparrow$ & AQ \scriptsize$\downarrow$ & EIG \scriptsize$\uparrow$ \\ 
\toprule
        & zero-shot  & 56.7\%          & 7.1               & 0.34 \\
INLG    & FT         & 46.6\%          & \textbf{4.6}      & 0.41 \\
        & DPO        & \textbf{68.9\%} & 5.2               & \textbf{0.45} \\ 
\midrule
        & zero-shot  & 51.1\%          & 7.5               & 0.29 \\ 
Things  & FT         & 42.2\%          & 5.4               & 0.31 \\ 
        & DPO        & \textbf{61.1\%} & \textbf{5.2}      & \textbf{0.40} \\ 
\midrule
        & zero-shot  & 71.1\%          & 7.6               & 0.35  \\
Celebrities & FT     & 46.7\%          & 5.5               & 0.39 \\ 
        & DPO        & 72.2\%          & \textbf{5.1}      & \textbf{0.47} \\
\bottomrule
\end{tabular}}
\caption{\textbf{Different Domains}: Results on candidate sets with $|\Omega|=8$. Across the three sets, DPO identifies the target within the first dialogue (S@1) more often, with shorter dialogues (AQ) and more informative questions (EIG). FT and DPO's improvements for AQ and EIG are statistically significant compared to zero-shot scores (Mann-Whitney U test).}
\label{tab:8cds}
\end{table}

Given the promising results of DPO, we proceed to evaluate the robustness of this approach with
larger candidate sets ($> 8$) at test time – preference optimization is still based on candidate sets of 8 elements.
As shown in Table~\ref{tab:bigger}, with candidate sets of 16 and 29 candidates, DPO outperforms zero-shot in terms of S@1, with improvements of $+ 6.7\%$ and $+13.8\%$, respectively. DPO generates shorter dialogues, reducing questions by $3.2$ and $0.7$. However, the average EIG of the DPO's questions slightly decreases compared to zero-shot in BigBench. 

\begin{table}[h]
\centering
\begin{tabular}{llccc} 
\toprule
Set & Method & S@1 \scriptsize$\uparrow$ & AQ \scriptsize$\downarrow$ & EIG \scriptsize$\uparrow$ \\ 
\toprule
INLG 16 & zero-shot & 44.4\% & 9.5 & 0.31 \\ 
& DPO & \textbf{51.1\%} & \textbf{6.3} & \textbf{0.38} \\
\midrule
BigBench 
& zero-shot & 17.2\% & 8.8 & 0.31 \\ 
& DPO & \textbf{31.0\%} & 8.1  & 0.28 \\ 
\bottomrule
\end{tabular}
\caption{\textbf{Different Size}: DPO boosts the model performance on INLG 16, and improves its success rate but not the other scores with BigBench. Improvements in INLG 16 for AQ and EIG are statistically significant (Mann-Whitney U test).}\label{tab:bigger}
\end{table}





\section{Analysis}


First of all, following~\citet{INLG23}, we examine the types of questions that models ask:  \emph{constraint seeking}  (CS) are questions about a feature shared by more candidates, whereas \emph{hypothesis scanning}  (HS) are questions about one single candidate.  In every test set apart from BigBench, zero-shot asks more CS  with lower informativeness than DPO (Table \ref{tab:typeQs}: EIG for CS questions has a $+0.17$ compared to zero-shot). In every test set apart from BigBench, we further observe DPO asking more HS questions than the zero-shot. Overall, DPO seems to follow a more effective information-seeking strategy: DPO narrows down to the relevant candidate subset by initially posing a series of CS questions, subsequently moving to mostly coherent HS. This holds also with abstract concepts, unseen during sampling, as illustrated in Figure~\ref{fig:example}.


\begin{table}[h]
\small
\centering
\scalebox{1.00}{
\begin{tabularx}{\columnwidth}{l l *{4}{>{\centering\arraybackslash}X}}
\toprule
Set & Method & \multicolumn{2}{c}{HS} & \multicolumn{2}{c}{CS} \\
\cmidrule(lr){3-4} \cmidrule(lr){5-6}
& & \% & EIG & \% & EIG \\
\toprule
INLG & zero-shot & 47.45 & 0.25 & 52.55 & 0.42 \\
 & DPO & 60.05 & 0.33 & 39.95 & \textbf{0.62} \\
\midrule
Things & zero-shot & 44.13 & 0.22 & 55.87 & 0.35 \\
 & DPO & 67.59 & 0.33 & 32.41 & \textbf{0.56} \\
\midrule
Celebrities & zero-shot & 31.84 & 0.15 & 68.16 & 0.44 \\
 & DPO & 53.92 & 0.36 & 46.08 & \textbf{0.59} \\
\midrule
INLG 16 & zero-shot & 38.66 & 0.19 & 61.34 & 0.39 \\
 & DPO & 51.47 & 0.27 & 48.53 & \textbf{0.51} \\
\midrule
BigBench & zero-shot & 64.2 & 0.22 & 35.71 & 0.48 \\
 & DPO & 61.64 & 0.21 & 38.36 & 0.41 \\
\bottomrule
\end{tabularx}
}
\caption{Percentage (\%) and average EIG for hypothesis scanning (HS) and constraint-seeking (CS) questions.}
\label{tab:typeQs}
\end{table}

\begin{figure}[h!]
\centering
\begin{tcolorbox}[colback=white, colframe=black, boxrule=0.1mm, arc=5mm, width=1.0\linewidth]
    \textbf{\small Target:} \small anger \\
    \small
    \textbf{List of candidates}:
    \inlinecolorbox{pastelYellow}{car}, 
    \inlinecolorbox{beaublue}{glass micropipette}, 
    global poverty, 
    \inlinecolorbox{pastelGrey}{house}, 
    \inlinecolorbox{bluebell}{anger}, 
    \inlinecolorbox{pastelYellow}{train}, 
    \inlinecolorbox{pastelYellow}{airplane}, 
    \inlinecolorbox{beaublue}{television}, 
    representative democracy, 
    \inlinecolorbox{bluebell}{love}, 
    \inlinecolorbox{celadon}{giganotosaurus}, 
    \inlinecolorbox{bluebell}{hate}, 
    tree, 
    \inlinecolorbox{bluebell}{surprise}, 
    \inlinecolorbox{bluebell}{disgust}, 
    \inlinecolorbox{celadon}{siberian husky}, 
    phase transition, 
    \inlinecolorbox{bubblegum}{shoes}, 
    frisbee, 
    \inlinecolorbox{bluebell}{jealousy}, 
    positive sum game, 
    beauty, 
    \inlinecolorbox{bubblegum}{coat}, 
    apple, 
    \inlinecolorbox{celadon}{cow}, 
    \inlinecolorbox{bluebell}{contentment}, 
    \inlinecolorbox{bluebell}{hopefulness}, 
    \inlinecolorbox{celadon}{dinosaur}, 
    \inlinecolorbox{pastelGrey}{shower}
    \tcblower
    \textbf{\small Dialogue}
    \vspace{0.5em} \\
    \small
    \textbf{Q:} Is the target an \inlinecolorbox{celadon}{animal}? \hfill \textbf{A:} No \\
    \small
    \textbf{Q:} Is the target a type of \inlinecolorbox{pastelYellow}{vehicle} (car, train, airplane, etc.)? \hfill \textbf{A:} No. \\
    \small
    \textbf{Q:} Is the target a type of \inlinecolorbox{pastelGrey}{building or structure} (house, train station, airport, etc.)? \hfill \textbf{A:} No. \\
    \small
    \textbf{Q:} Is the target a form of \inlinecolorbox{beaublue}{technology} (computer, phone, micropipette, etc.)? \hfill \textbf{A:} No. \\
    \small
    \textbf{Q:} Is the target a type of \inlinecolorbox{bubblegum}{clothing or accessory} (shirt, pants, hat, etc.)? \hfill \textbf{A:} No. \\
    \small
    \textbf{Q:} Is the target a type of \inlinecolorbox{bluebell}{emotion or feeling} (love, hate, joy, etc.)? \hfill \textbf{A:} Yes. \\
    \small
    \textbf{Q:} Great! Is the target \lq\lq \textbf{love}\rq\rq? \hfill \textbf{A:} No. \\
    \small
    \textbf{Q:} Is the target \lq\lq\textbf{hate}\rq\rq? \hfill \textbf{A:} No. \\
    ... \\
    \small
    \textbf{Q:} Is the target \lq\lq \textbf{anger}\rq\rq? \hfill \textbf{A:} Yes.
\end{tcolorbox}
\caption{Example from the BigBench. DPO asks grounded CS questions (highlighted with colors), identifying the subset of the target (e.g., emotions). It then asks a series of HS questions.}
\label{fig:example}
\end{figure}


Secondly, we aim to understand the high scores DPO obtains in INLG and the lower ones in BigBench. INLG candidate sets are organized taxonomically, half of the candidates belong to one category and the other half to another one. By inspecting the dialogues, we saw that both zero-shot and DPO tend to ask informative questions at the first turn, identifying one of the two categories. However, the two models differ in the follow-up turn: when the first turn receives a negative answer, zero-shot questions tend to have low EIG, while DPO questions are informative by zooming on the identified category (more on negation in Appendix~\ref{appendix:analysis}). This suggests that DPO has learned to profit from negatively answered questions, on which LLMs are known to fail~\cite{NegationCondaQa,truong-etal-2023-language}. We dive into BigBench by comparing the models' performance on concrete and abstract targets (Table~\ref{tab:bigben}). In general, both zero-shot and DPO perform better when the target is a concrete entity. Notably, DPO significantly improves performance with both types of concepts (+12.5\% and +15.4\% for concrete and abstract targets, respectively). This improvement, however, is not reflected in the average dialogue length when the target is abstract. The significant difference in guessing concrete vs. abstract targets (25.0\% vs. 7.7\% for zero-shot and 37.5\% vs. 23.1\% for DPO) calls for further investigation into the underlying factors contributing to this disparity and the potential need for tailored strategies to handle abstract concepts more effectively.

\begin{table}[h]
\centering
\begin{tabular}{llccc} 
\toprule
Set & Method & S@1 \scriptsize$\uparrow$ & AQ \scriptsize$\downarrow$ & EIG \scriptsize$\uparrow$ \\ 
\toprule
Concrete & zero-shot & 25.0\% & 8.4 & 0.33 \\
& DPO & \textbf{37.5\%} & \textbf{6.3} & \textbf{0.34} \\ 
\midrule
Abstract 
& zero-shot & 7.7\% & 9.3 &  0.29 \\ 
& DPO & \textbf{23.1\%} & 10.3  & 0.24 \\ 
\bottomrule
\end{tabular}
\caption{BigBench: concrete vs. abstract target.}\label{tab:bigben}
\end{table}

%% file: latex/conclusion.tex
\section{Conclusion}

In our work, we designed a Direct Preference Optimization (DPO) approach to enhance the informativeness of questions asked by LLMs using Expected Information Gain (EIG). We utilized the 20 Questions Game paradigm, a framework in cognitive science and AI for studying information-seeking behavior, reasoning, and hypothesis testing. Our approach involved (a) sampling multiple questions from the model in a zero-shot fashion, (b) clustering the questions based on their EIG, and (c) training the model using these clusters with DPO. Our results show that DPO significantly improves question informativeness (measured by average EIG and number of questions asked), making the dialogue strategy more effective. Notably, this method generalizes well to different domains. Our findings demonstrate that EIG is a promising training signal for improving the reasoning capabilities of LLMs in information-seeking dialogues.

%% file: latex/limitations.tex
\section{Limitations}
This work is intended to be exploratory. Our limitations pertain to three main categories: the controlled paradigm employed, the model and the training regime tested, and the EIG computation. 

Our version of the 20 Questions game poses two limits: a closed set of possible candidates, and polar questions and answers. We consider it necessary to investigate this controlled setting before transitioning to more realistic scenarios. Nevertheless, possible solutions to compute EIG in an open setting emerge from the literature: \citet{hu:unce24} rely on the expected candidates from the model; \citet{Apple}, instead, build the set of initial candidates from external sources. Transitioning from polar questions to more realistic scenarios remains an open challenge. Related studies, however, already provide useful strategies to overcome this restriction, computing EIG with different-shaped \citep{grand:LIPS} and open-ended questions \citep{GroundedRetrieval23}. 

Secondly, to investigate whether LLMs could learn to be more informative and effective with self-generated EIG signals, we focus on one model (\llm{Llama 2-chat 7B}) and one preference optimization strategy (DPO). We select \llm{Llama 2-chat 7B} as the best-performing open-source model at the start of the project. According to our computing resources (2X24 GB GPUs), larger 13B and 70B models’ versions could not be trained without quantization. In preliminary studies on the 7B and 13B versions (zero-shot), we observed quantization (8/4 bit) leading to performance degradation–on the S@1 metric in particular. For this reason, we preferred to focus on the 7B version. Further work is required to determine if this training strategy holds with other models and other preference optimization strategies. 

A third limitation of our study is related to the EIG computation. EIG computation depends on the yes/no annotation. While we can assume a high degree of accuracy, based on our careful inspection of the dialogues, some questions are still difficult to answer with only yes/no (e.g., $w =$ dinosaur, the question \lq\lq Is it living?\rq\rq). Additionally, all the candidates in $\Omega$ are considered equally likely, while LLMs have priors conditioning their question generation. Furthermore, when computing the EIG of follow-up questions, we consider the model able to sequentially rule out candidates excluded in the dialogue history, which could be a strong assumption for a generative language model. 

%% file: latex/appendix.tex
\section{Methodology} \label{sec:methodology}

\subsection{Successful Condition} \label{appendix:successful_cond}

A dialogue is \emph{successful} when the Questioner guesses the target entity. We have implemented this condition with the following two cases:  (a) the Questioner explicitly refers to the target entity without naming other candidate entities or
(b) the Questioner explicitly refers to the target entity along with other candidates, and the target is named more times than each of the other entity. This second condition recognizes when the Questioner is explicitly reasoning about the target before guessing (e.g., target: \lq cherry\rq. Questioner: \lq So, the target is one of the remaining fruits, which are cherry and grapefruit. Let me make a guess. Is the target... cherry?\rq). The dialogue is considered \emph{unsuccessful} if the Questioner does not meet the conclusive conditions above within 20 questions. 

\subsection{Expected Information Gain}\label{appendix:eig}

In this setting, each question aims to rule out as many candidates as possible from the set $\Omega = \{\omega_{1}, \omega_{2}, ..., \omega_{n}\}$. The most informative question rules out half of the possible candidates, whereas a question ruling out none or just one candidate has low informativeness. Questions' informativeness has been quantified in prior works through the Expected Information Gain.

Expected Information Gain (EIG) measures the contribute of each question to reduce uncertainty towards the solution $\omega$. The level of uncertainty is measured through the entropy $H$, higher entropy means higher uncertainty. Consequently, EIG is computed subtracting from the initial state of entropy ($H_{prior}$) the expected entropy after asking the question ($H_{posterior}$).

\begin{equation}\label{eq:1}
    EIG  = H_{prior} - H_{posterior}
\end{equation}


At the beginning of each game, we assume a uniform prior distribution over all the possible candidates $\omega_{1}, \omega_{2}, ..., \omega_{n}$. From the Shannon entropy \citep{shannon1948mathematical}, this initial state $H_{prior}$ in our setting is equal to:

\begin{equation}\label{eq:2}
\begin{split}
    H_{prior} & = - \sum_{i=1}^n \: p(w_{i}) \: log_{2} \: p(w_{i}) \\
    & =  - n \: \frac{1}{n} \: log_{2} \: \frac{1}{n} \: = \: log_{2} \: {n}
\end{split}
\end{equation}

A yes/no question divides the current candidates in $\Omega$ into two disjoint subsets: $\Omega_{yes} \,  \cup \, \Omega_{no}$. In the subset $\Omega_{yes}$ there are the candidates in $\Omega$ that are consistent with a possible positive answer, $\Omega_{no}$ with a negative one. Answering the question, the Answerer determines which subset becomes the new set of candidates under consideration for the next turn (the candidates in the other subset are ruled out). The posterior entropy is computed as the sum of the entropies of both the yes and no subsets, weighted by their probability. 
$H_{posterior}$ is computed as: 

\begin{equation} \label{eq:3}
    H_{posterior}  = p_{yes} \, H_{yes} \: + \: p_{no} \, H_{no}
\end{equation}

This computation is based on the expected answers to the question for each candidate in the current set. In other words, the Questioner has an expectation of which candidates will be in $\Omega_{yes}$ and $\Omega_{no}$ before receiving the answer by the Answerer. 
With these subsets, the probabilities $p_{yes}$ and $p_{no}$ refer to the probabilities of receiving a positive or negative answer. The entropies $H_{yes}$ and $H_{no}$ measure uncertainty in the two subsets computed as in eq. \ref{eq:2}. Intuitively, the higher the similarity between the two subsets in terms of size (i.e., candidates per subset) the higher the EIG value.
The most informative yes/no questions divide $\Omega$ into two subsets $\Omega_{yes}$ and $\Omega_{no}$ of the same size, resulting in $EIG = 1$. 




\subsection{Setting} \label{appendix:setting}

\subsubsection{Sampling}

\begin{figure*}[htbp]
\begin{center}
  \includegraphics[width=1.0\textwidth]{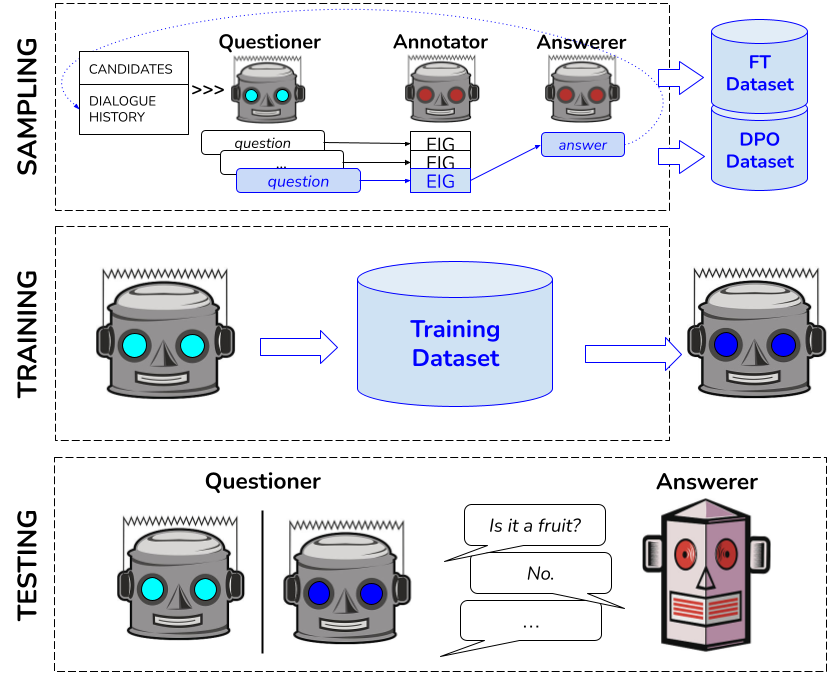}
  \caption{During \textbf{sampling}, \llm{Llama 2-chat (7B)} plays the roles of the Questioner, the Annotator, and the Answerer. Questions are sampled from the Questioner and then evaluated by the Annotator. Once an optimal question is reached, the Answerer answers it. The optimal question and its answer are appended to the dialogue history. In this way, optimal questions are sampled not only for the first turn but also in follow-up turns. In \textbf{training}, the Questioner is trained with FT and DPO datasets. In \textbf{testing}, the zero-shot \includegraphics[height=1.0em]{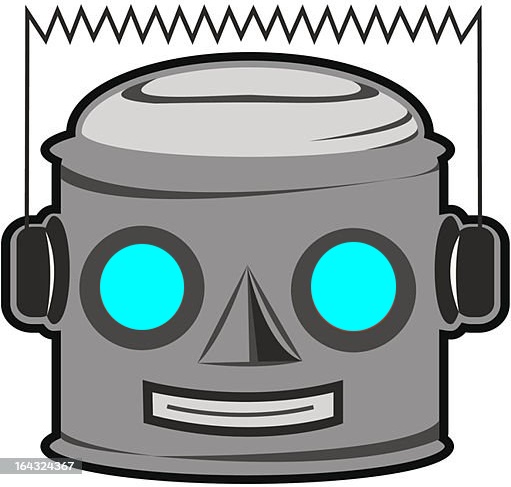} and trained \includegraphics[height=1.0em]{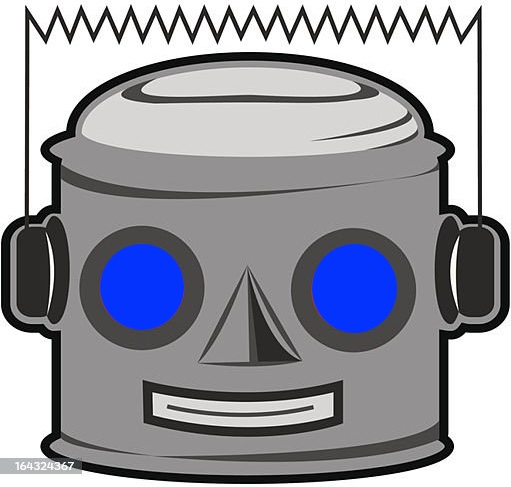} Questioners play the 20 Questions Game with an external model as Answerer.}
  \label{fig:process_figure}
\end{center}
\end{figure*}

Our approach comprises the three steps described in Fig. \ref{fig:process_figure}. 
During sampling, different questions are sampled from the Questioner \includegraphics[height=1.5\fontcharht\font`\B]{images/IP_4.jpg}. To compute EIG, the Annotator \includegraphics[height=1.5\fontcharht\font`\B]{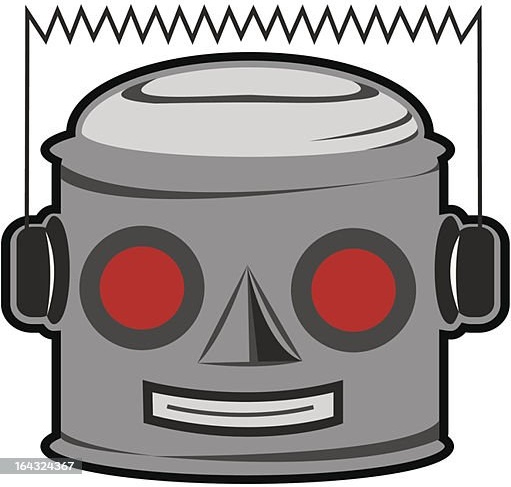} annotates the questions with yes/no for each candidate in the candidate set ($\Omega$). Once an \lq optimal\rq question is reached ($EIG=1$), the optimal and suboptimal questions are stored in the training datasets. All the prompts for the roles are reported below. The Annotator has the same prompt as the Answerer, taking as a target each candidate in the candidate set.

\begin{tcolorbox}[colback=pastelGrey,colframe=gray,width=\columnwidth,arc=5mm,auto outer arc]
  \begin{center}
    \textbf{Questioner} 
    \includegraphics[height=2.0\fontcharht\font`\B]{images/IP_4.jpg}
    \includegraphics[height=2.0\fontcharht\font`\B]{images/IP_3.jpg}
  \end{center} 
  
    \textbf{System Prompt:} \emph{You are playing a game, make only one yes/no question at turn to identify the target from the List of candidates. If there are 1 or 2 candidates remaining make the guess.} 
    
    \textbf{User Prompt:} \emph{List of candidates:  \inlinecolorbox{white}{\texttt{\small{CANDIDATES}}}}.
\end{tcolorbox}

\begin{tcolorbox}[colback=pastelGrey,colframe=gray,width=\columnwidth,arc=5mm,auto outer arc]
  \begin{center}
    \textbf{Answerer/Annotator}
    \includegraphics[height=2.0\fontcharht\font`\B]{images/IP_2.jpg}
    \includegraphics[height=2.0\fontcharht\font`\B]{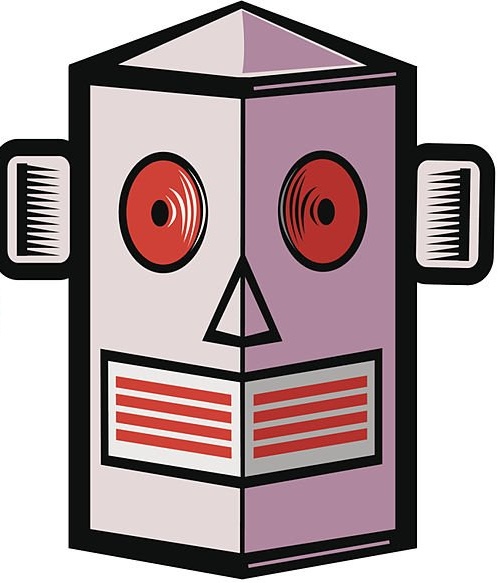}
  \end{center} 
  
    \textbf{System Prompt:} \emph{You are playing the 20-Questions game, you will be asked one Question about the Target element. Answer only \lq yes' or \lq no' to the Question depending on your Target element.}
    
    \textbf{User Prompt:} \emph{Target element: 
    \inlinecolorbox{white}{\texttt{\small{TARGET}}}, \\ Question:
    \inlinecolorbox{white}{\texttt{\small{QUESTION}}}} 
\end{tcolorbox}

To yield diverse outputs during sampling, the decoding strategy for the Questioner involves a relatively high temperature (1.0) and top-k sampling (50). To achieve more deterministic responses the Answerer/Annotator has a low temperature of 0.1.

This process produces a Fine-Tuning and DPO dataset. The Fine-Tuning dataset consists of all the completed dialogues of optimal questions only. The DPO dataset, instead, comprises pairs of \lq optimal\rq questions ($EIG = 1.0$) with suboptimal questions ($EIG < 0.8$). 

\subsubsection{Training}
The Fine-Tuning and DPO datasets are employed to train \llm{Llama 2-chat (7B)} with LoRA adapters \citep{LloraAdapters}. Our main focus is on the approach rather than the training strategy, thus we did not perform extensive hyperparameter tuning. We do not exclude that additional experiments on hyperparameters could lead to better results for our approach. For both Fine-Tuning and DPO, the LoRA adapters are applied on all the modules, with an $r=128$ and the alpha value $=32$. We perform Fine-Tuning and DPO with AdamW as the optimizer, training for one epoch and with a batch size $= 8$. For both Fine-Tuning and DPO, a learning rate of $2e-5$ has been observed to produce consistent results.

\subsubsection{Testing}
At test time, we compare the \llm{Llama 2-chat (7B)} zero-shot with its Fine-Tuned and DPO alternatives. We employ an external model as Answerer to make the test setting more different from the training setting. The Answerer model \includegraphics[height=1.5\fontcharht\font`\B]{images/IP_1.jpg} is \llm{gpt-3.5-turbo-0125}
, with default parameters for generation. The total cost for testing - every approach in every setting - is less than 3\$. At test time, we experimented also with the Questioner using the default \llm{Llama 2-chat} hyperparameters, the results are consistent with the ones reported in the paper Section~\ref{sec:results}. For the hyperparameters used in the paper, the results across multiple runs are also consistent.

\subsubsection*{Implementation Details}
The inference and training with \llm{Llama 2-chat (7B)} were performed using the Huggingface Transformer library \citep{HuggingFace}. All experiments were executed on a Nvidia GeForce RTX 3090 24GB GPU. Sampling 10,000 dialogues took approximately 4 days. Fine-Tuning and DPO take around 12 hours each. Running inference on all the test sets lasts around 6 hours for FT and DPO, whereas it lasts around 10 hours for the zero-shot model.

\section{Test Sets} \label{sec:apdataset}

The games in the test sets are with candidates unseen during training. To test the trained models in different domains, we have the following test sets of 8 candidates, as in training:

\begin{itemize}

    \item \textbf{INLG}: 90 
    candidate sets structured in such a way that half of the candidates pertain to one taxonomic category and the other half to another one \citep{INLG23}.
    The taxonomic categories for the candidates are animals, clothing, weapons, fruits, and vegetables. 
    
    \item \textbf{Things}: 90 candidate sets from the evaluation split of \citet{Apple}. The full list of categories are animals, clothing, foods, objects, plants, vehicles,  professions, materials, instruments, places, sports, buildings, furniture, celestial bodies, mythical creatures, events, and activities. 

    \item \textbf{Celebrities}: 90 sets from the evaluation split of \citet{Apple}. In this test set, the candidates differ greatly from the training ones, comprising only past and living celebrities.

\end{itemize}

\noindent To test the trained models with more candidates ($\Omega > 8$), we have the following test sets:

\begin{itemize}
    \item \textbf{INLG 16}: 90 candidate sets of 16 structured elements, as in INLG \citep{INLG23}.
    \item \textbf{BigBench}: 29 sets of 29 candidates of abstract and concrete concepts. The concepts are apple, television, dinosaur, airplane, house, tree, coat, shoes, car, train, shower, frisbee, cow, giganotosaurus, siberian husky, glass micropipette (concrete); anger, love, hate, contentment, jealousy, surprise, disgust, hopefulness, global poverty, phase transition, positive sum game, beauty, representative democracy (abstract). 
\end{itemize}

\section{Analysis} \label{appendix:analysis}

\subsection*{Informativeness after a Negative Answer}

INLG (8) and INLG 16 consist of structured sets, i.e., half of the candidates of one category and the other half to another one. 
Both the zero-shot and the DPO tend to ask informative questions at the first turn, identifying one of the two categories. After a negative answer to this first question, the zero-shot frequently asks a confirmation question about the other category ($EIG=0$), as illustrated in Figure \ref{fig:example_neg}. 
In both INLG 8 and INLG 16 of Table \ref{tab:yesno}, the zero-shot shows a higher percentage of 0-EIG questions and lower average EIG of questions after a negative answer, compared to questions after a positive answer (EIG after negative and positive questions drawn from different distributions, Mann–Whitney U test). On the contrary, the DPO has a lower percentage of 0-EIG questions and higher EIG after a negative answer (EIG after negative and positive questions of the same distribution). Trained on similarly structured candidate sets, the DPO seems not to reproduce the uninformative behaviour of the zero-shot, asking more informative questions after negative answers.


\begin{table}[htbp]
\centering
\small
\setlength{\tabcolsep}{3pt}
\begin{adjustbox}{max width=\textwidth}
\begin{tabular}{@{}lp{1.5cm}cccc@{}}
\toprule
Set & Method & \multicolumn{2}{c}{\textbf{after \emph{yes}}} & \multicolumn{2}{c}{\textbf{after \emph{no}}} \\
\cmidrule(lr){3-4} \cmidrule(lr){5-6}
& & EIG $\uparrow$ & \%0 EIG $\downarrow$ & EIG $\uparrow$ & \%0 EIG $\downarrow$ \\
\midrule
INLG 8 & zero-shot & 0.36 & 60.1\% & 0.21 & 74.1\% \\
       & DPO       & 0.37 & 59.4\% & \textbf{0.33} & \textbf{61.6\%} \\
\midrule
INLG 16 & zero-shot & 0.35 & 55.6\% & 0.20 & 71.5\% \\
        & DPO       & 0.29 & 65.6\% & \textbf{0.29} & \textbf{60.2\%} \\
\bottomrule
\end{tabular}
\end{adjustbox}
\caption{Questions' informativeness after a positive and a negative answer, in terms of average EIG (EIG) and percentage of uninformative questions (\% 0 EIG).}
\label{tab:yesno}
\end{table}

\begin{figure*}[htbp]
\centering
\tcbset{
  width=\textwidth, 
  colframe=black, 
  colback=white, 
  boxrule=0.1mm, 
  arc=5mm, 
  before skip=0mm, 
  colbacktitle=white, 
  fonttitle=\bfseries, 
  center title
}
\begin{tcolorbox}
\textbf{Target:} fox \\ 
\textbf{List of candidates}: \emph{\inlinecolorbox{beaublue}{elk}}, chicken, robin, starling, \inlinecolorbox{beaublue}{fox}, partridge, \inlinecolorbox{beaublue}{hamster}, \emph{\inlinecolorbox{beaublue}{buffalo}}
\tcblower
\begin{minipage}[t]{0.48\textwidth} 
\raggedright
\vspace{0pt} 
\textbf{Zero-shot} \\
\vspace{0.2em} 
\textbf{Q$_1$:} Is the target a bird? \textsuperscript{\textcolor{pastelGreen}{$EIG=1$}} \hfill \textbf{A: No.} \\
\textbf{Q$_2$:} Is the target a \inlinecolorbox{beaublue}{mammal}? \textsuperscript{\textcolor{red}{$EIG=0$}} \hfill \textbf{A:} Yes. \\
\textbf{Q$_3$:} Does the target have a long tail? \hfill \textbf{A:} Yes. \\
\textbf{Q$_4$:} Is the target a fox? \hfill \textbf{A:} Yes.
\end{minipage}%
\hfill
\vrule
\hfill
\begin{minipage}[t]{0.48\textwidth}
\raggedright
\vspace{0pt} 
\textbf{DPO} \\
\vspace{0.2em} 
\textbf{Q$_1$:} Is the target bird? \textsuperscript{\textcolor{pastelGreen}{$EIG=1$}} \hfill \textbf{A: No.} \\
\textbf{Q$_2$:} Is the target \emph{\inlinecolorbox{beaublue}{mammal a large hoofed}} animal, like a deer or a bison? \textsuperscript{\textcolor{pastelGreen}{$EIG=1$}} \hfill \textbf{A:} No. \\
\textbf{Q$_3$:} Is the target a carnivorous mammal, like a fox? \\[-0.3em] \hfill \textbf{A:} Yes.
\end{minipage}
\end{tcolorbox}
\caption{INLG 8 Example: half of the candidates are birds, and half are mammals. Both the zero-shot and DPO identify the bird category at the first turn ($EIG=1$). After the negative answer, the zero-shot asks a confirmation question about the remaining category (i.e., mammal), with an $EIG=0$. The DPO, instead, asks a more specific question on the remaining category (i.e., large hoofed, which are the \lq elk\rq and \lq buffalo\rq), with an $EIG=1$. The higher informativeness of DPO is further reflected in the number of questions required to reach the target: 3 questions in DPO vs. 4 questions in zero-shot.}
\label{fig:example_neg}
\end{figure*}